# FV2ES: A Fully End2End Multimodal System for Fast Yet Effective Video Emotion Recognition Inference

Qinglan Wei, Xuling Huang, Yuan Zhang

*Abstract*—In the latest social networks, more and more people prefer to express their emotions in videos through text, speech, and rich facial expressions. Multimodal video emotion analysis techniques can help understand users' inner world automatically based on human expressions and gestures in images, tones in voices, and recognized natural language. However, in the existing research, the acoustic modality has long been in a marginal position as compared to visual and textual modalities. That is, it tends to be more difficult to improve the contribution of the acoustic modality for the whole multimodal emotion recognition task. Besides, although better performance can be obtained by introducing common deep learning methods, the complex structures of these training models always result in low inference efficiency, especially when exposed to high-resolution and long-length videos. Moreover, the lack of a fully end-to-end multimodal video emotion recognition system hinders its application. In this paper, we designed a fully multimodal video-to-emotion system (named FV2ES) for fast yet effective recognition inference, whose benefits are threefold: (1) The adoption of the hierarchical attention method upon the sound spectra breaks through the limited contribution of the acoustic modality, and outperforms the existing models' performance on both IEMOCAP and CMU-MOSEI datasets; (2) the introduction of the idea of multi-scale for visual extraction while single-branch for inference brings higher efficiency and maintains the prediction accuracy at the same time; (3) the further integration of data pre-processing into the aligned multimodal learning model allows the significant reduction of computational costs and storage space. Source code will be available in https://github.com/qlwei89/FV2ES if this paper is accepted.

## I. Introduction

WITH the rapid development of social networks, more people choose to use videos to express their emotions and opinions. The modern technique of multimodal emotion recognition can help understand these emotions through text, speech, facial expressions, gestures, postures, etc. In the existing multimodal-based video emotion recognition work, we observe three distinct limitations. **Firstly**, acoustic modality features are usually extracted by the OpenSmile toolkit [1] or RNN-based deep learning networks [2, 3, 4]. Though these global spectrum features or coarse-grained audio information are useful for the overall multimodal recognition task, it seems that the contribution of the acoustic modality is relatively low as compared to visual and textual modalities, which affects the improvement of the video emotion recognition performance. **Secondly**, 5G networks allow people to easily record and share high-definition videos through their mobile phones anytime and anywhere. Meanwhile, in the era of self-media, the low threshold of video lease has significantly increased the number and the length of videos. However, although recent deep learning networks [5, 6, 7, 8, 9] adopted for the visual modality led to better performance, they cannot cope with the above-mentioned computing and storage challenges due to the complex structures. **Thirdly**, throughout the entire field of multimodal video emotion recognition, the research has always stayed at the academic level. Most existing non-end-to-end frameworks [10, 11, 12] own pipelines with combined models that aim to obtain better performance; however, this kind of slow and incoherent method hindered the application of multimodal video emotion recognition.

In this paper, we aim to address the above limitations by designing a fully multimodal video-to-emotion system for fast yet effective recognition inference, whose main flow and components are illustrated in Fig. 1. In this system, the original video is taken as the input. The method of hierarchical attention is used to extract features for each spectral patch of the audio modality to enhance the effect of the audio modality. At the same time, we adopt multi-branch feature learning and single-branch inference structure, so that we can extract the frames' information of visual modality and simplify the inference model. For text modalities, we adopt Albert [13] to extract textual features. And we use the basic transformer [14] to obtain the visual and acoustic sequential information. Finally, multimodal fusion is performed through a feedforward network.

Specifically, through sound spectrum segmentation, intra-block self-attention, and block aggregation processing, we extract the layered spectral features to obtain the internal relationship information of the audio spectrum. At the same time, for video frames, we adopt the structure of multi-branch feature learning and single-branch inference. During training, the advantages of multi-branch feature learning are used to extract the information of video frames. In the inference process, we choose the single-branch structure for prediction, which is simpler and more computationally efficient. We take their respective strengths and realize their connection through parameter migration, which can compress the computational loss and ensure the learning of frame features. Simultaneously, to realize the comprehensive connection between the input video and the procedure of emotion analysis, we integrate the data pre-processing into the multimodal emotion analysis model.

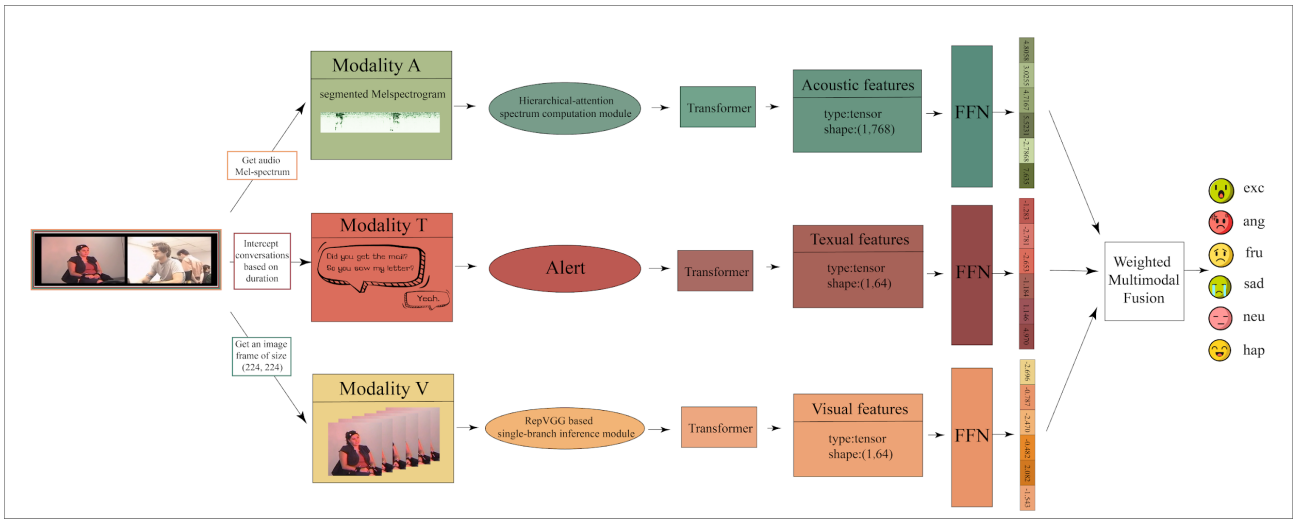

Fig. 1. The data flow of the proposed V2EM.

Our model improves prediction efficiency and promotes the feasibility of industrial application. Experimental results show that the processing of visual and acoustic perception enables our model to significantly outperform the existing multimodal emotion analysis models.

**The main contributions and innovations of this paper include:**

- We successfully migrate the success of Transformer in Vision (ViT) to the audio modality. The hierarchical audio spectrum information breaks through the contribution limitation and effectively improves the performances on two public datasets.
- The idea of multi-scale visual feature extraction and single-branch reasoning is introduced to ensure the prediction accuracy and improve the efficiency of multimodal emotion analysis.
- The multimodal learning model that integrates video pre-processing and emotion analysis greatly reduces the computational cost and storage space, making it industrially applicable.

## II. RELATED WORK

Over the past few years, multimodal emotion recognition has been a widely studied academic topic for various research tasks such as human-machine communication [30], action recognition [31], online advertising recommendation [32] and Video Q&A [33,34]. In Table I, we show almost all of the existing popular multimodal emotion recognition research. **From the "Modality" column**, we can find that all of the multimodal emotion recognition methods considered both visual (V) and textual (T) modalities. It is worth noting that in some studies, researchers studied only V and T modalities while ignoring the acoustic (A) modality [7, 9]. "Effect" in this table refers to the performance effect of three modalities on common datasets and evaluation standards, and it seems that the contribution of the acoustic modality was always the lowest. This kind of low contribution mostly embodies two aspects. **On one hand**, the emotion recognition performance of modality A is worst compared to modality V and T. For example, in HFU-BERT [6], VGGish & SoundNet were used to extract audio information and the acoustic modality played the auxiliary role (F1-scores on the CMU-MOSEI dataset equal 0.74, 0.72, 0.56 for modality V, T, A respectively). **On the other hand**, it seems more difficult to improve the performance of modality A in multimodal emotion recognition research. For instance, the ablation experiments in [22] show that accuracies were increased by 7.4%, 9.6% while only 6.7% on the CMU-MOSEI dataset for modalities V, T, and A respectively.

In addition, **looking at the "Visual processing" column** in Table I, we can see that after the year 2015, almost all of the researchers tend to introduce deep learning networks (Cross-modal Transformer [23], ResNet50 [6]) to obtain deeper visual features for better multimodal emotion prediction effect. However, deeper networks own more complex structures, leading to lower computing efficiency for inference and more demands for storage memories. Thus, deep learning inference models may not function well in the current 5G and self-media era, where videos with higher resolution and longer length are leased and shared. This practical problem has yet to be addressed in the existing research.

Moreover, **the "System implementation" column shows** that there is still no attempt to build a multimodal video-to-emotion system to promote practical application, and that may be due to the non-end2end structures. Specifically, in [10], late fusion [11, 12], and later tensor fusion [35], manual feature learning or common pretrained deep learning models were usually combined [7, 23]. This can result in the unaligned fusion relationship among the V, T, and A modalities in the training and inference procedure. Though Dai W et al. [5] made a breakthrough by realizing end-to-end aligned multimodal features learning, this model needed the data pre-processing module and thus it still has not reached the full consistency from data input to emotion prediction.

## III. PROPOSED METHODS

The designed video-to-emotion multimodal model (named V2EM) for video emotion recognition inference is

TABLE I

SUMMARY OF EXISTING MULTIMODAL EMOTION RECOGNITION RESEARCH. THE TABLE SHOWS THE EFFECT OF THESE METHODS, THE CONTRIBUTION OF VARIOUS MODALITIES, HOW EACH STUDY PROCESSES VISUAL INFORMATION AND WHETHER A COMPLETE SYSTEM FOR MULTIMODAL EMOTION RECOGNITION IS PROVIDED.

| Years | Method | Modality | Effect | Visual processing | System implement |
|---|---|---|---|---|---|
| 1998 | Chen L S et al. [15] | V+A | V>A | HMM | ✗ |
| 2013 | Rosas V P et al. [16] | V+A+T | T>V>A | OKAO | ✗ |
| 2015 | Pang L et al. [17] | V+A | V>A | DBM | ✗ |
| 2016 | DemoFV [18] | V+A | V>A | DBN | ✗ |
| 2016 | CRMKL [19] | V+A+T | V>T>A | RNN+MKL | ✗ |
| 2016 | Seng K P et al. [20] | V+A | V>A | OKL-RBF | ✗ |
| 2017 | Nguyen D et al. [21] | V+A | V>A | C3D+DBN | ✗ |
| 2019 | Agarwal A et al. [22] | V+A+T | T>A>V | LRNN | ✗ |
| 2019 | MulT [23] | V+A+T | T>V>A | Crossmodal Transformer | ✗ |
| 2020 | MMLGAN [8] | V+A+T | V>A>T | ResNet | ✗ |
| 2020 | Dresvyanskiy D et al. [24] | V+A | V>A | VGGFER | ✗ |
| 2020 | ICCN [25] | V+A+T | T>V>A | DCCA | ✗ |
| 2021 | CFN-SR [26] | V+A | V>A | I3D | ✗ |
| 2021 | HFU-BERT [6] | V+A+T | V>T>A | VGG16+ResNet50 | ✗ |
| 2021 | Lin Li et al. [7] | V+T | Nolisted | VGG16 | ✗ |
| 2021 | FE2E [5] | V+A+T | T>V>A | VGG13 | ✗ |
| 2021 | ERLDK [27] | V+A+T | Nolisted | GRUCell | ✗ |
| 2022 | Praveen Rajasekar G et al. [28] | V+A | V>A | I3D+TCN | ✗ |
| 2022 | MTTCCT [29] | V+A+T | T>V>A | sLSTM | ✗ |
| 2022 | CMCN [9] | V+T | T>V | VGG16 | ✗ |

illustrated in Fig. 1. In this model, we input the original video including visual, textual and acoustic modalities. We use hierarchical attention to extract features for each spectral patch of the audio modality. And the RepVGG-based single-branch inference module is introduced to obtain the frame information of visual modality. For text modalities, we adopt Albert [13] to extract features. Then, the basic transformer [14] is used to get the visual and acoustic sequential information. Finally, we perform the weighted multimodal fusion with the feedforward network. In the following sections, we give the details of these three main technological innovations.

*A. The Hierarchical-Attention Spectrum Computing Module*

In order to bridge the gap between the contributions to the final emotion prediction results of the acoustic modality and the other two modalities, and inspired by the outstanding performance of nesting transformers [36] applied in image classification, we design a novel hierarchical-attention spectrum computing module to get fine-grained spectral information. Fig. 2 illustrates the structure of the module proposed in this section. The right panel of this figure looks like a three-layer pyramid, where spectrum maps at various layers are included. And the left panel shows the procedure of generating hierarchical spectrum maps.

Specifically, the initial spectrum is obtained through Mel-scale filter banks. The input of the hierarchical-attention spectrum computing module is a reshaped $H \times W$ sound spectrum, where $H = W$. And this input is divided into 16 patches of equal size $S \times S$ (where $S = \frac{H}{4} = \frac{W}{4}$), and these little patches are called the first-layer spectrum maps. After segmentation, we use the Transformer Layer for each spectral patch. Inside the Transformer Layer, we firstly embed the patch of size $S \times S$ as a $d$-dimensional vector:

$$I\ embedding \to x \in R^d \quad (1)$$

where $I$ refers to a patch size of $S \times S$.

Then, the basic transformer [14] is adopted on each $d$-dimensional vector to extract local self-attention acoustic features. And the LN [37] and GELU [38] operations are performed on the features to get the output of the first layer:

$$O_1 = GELU(LN(x + MSA(x))) \quad (2)$$

where $O_1$ represents the output of the first layer, $MSA$ is the multi-head attention of the basic transformer. As shown in Fig. 2, every four spatially adjacent patches are merged into one block through a $3 \times 3$ CNN followed by LayerNorm [37] and a $3 \times 3$ max pooling layer. The aggregated four blocks are termed as the second-layer spectrum maps:

$$I_2 = MaxPooling\left(LN\big(CNN(O_1)\big)\right) \quad (3)$$

where $I_2$ is the input of the second layer. With regard to the second layer, each block is again fed into the basic transformer [14] and then these four blocks are combined. The result is the input of the third layer $I_3$, which is passed through the basic transformer to get the final fused spectrum map.

Overall, we extract acoustic hierarchical spectral information through sound spectrum segmentation, intra-patch (block) self-attention, and patch (block) aggregation. In this way, we take full advantage of the internal relationship of the acoustic data, making up for the limitations of the previous work in which the global and coarse-grained acoustic information is extracted.

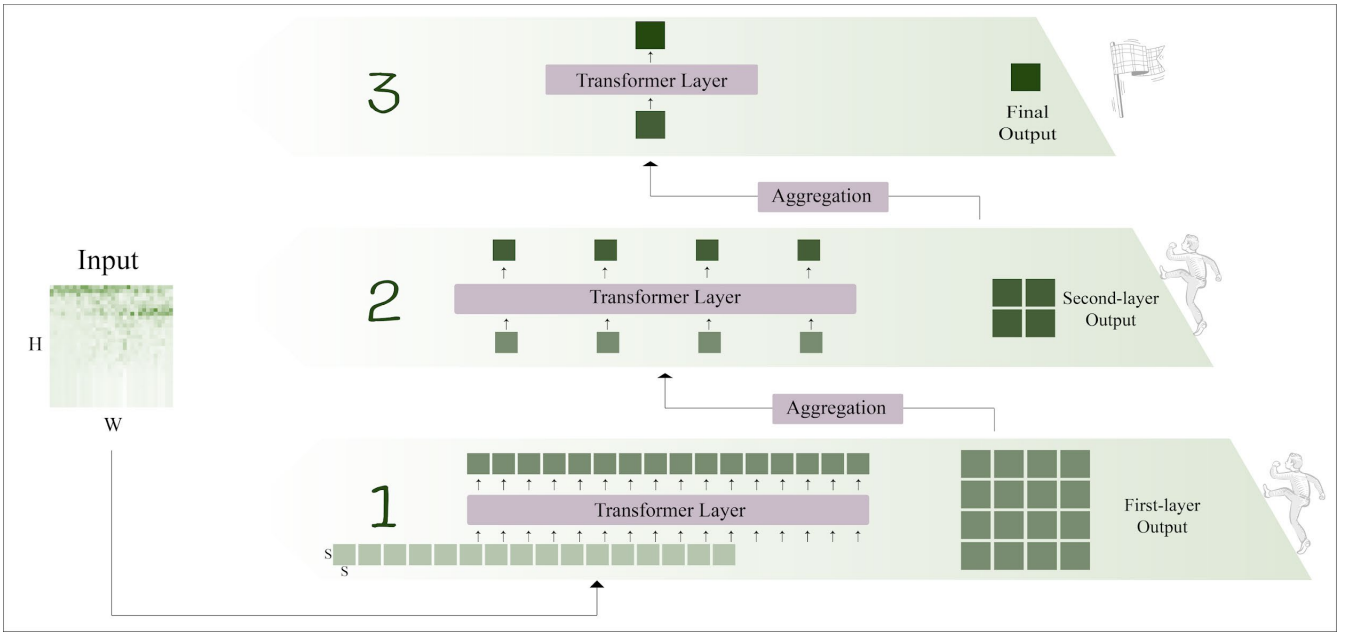

Fig. 2. The structure of the hierarchical-attention spectrum computing module.

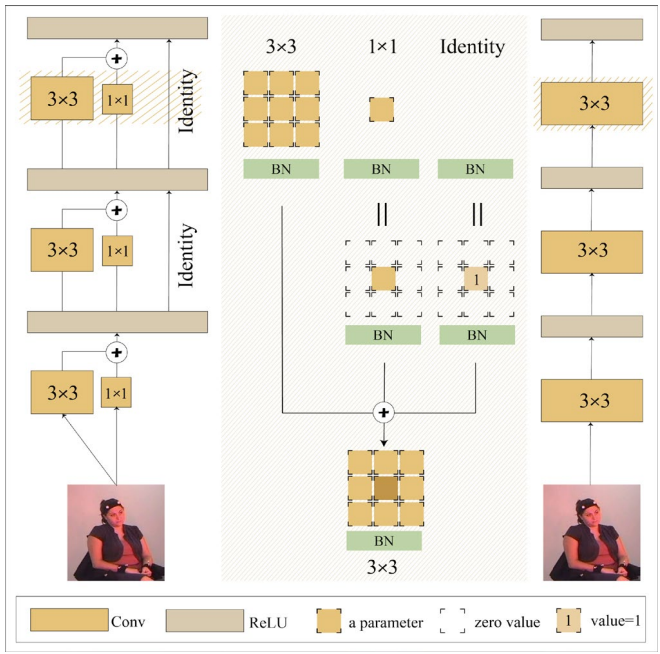

Fig. 3. Illustration of the method of the RepVGG-based single-branch inference module.

*B. The RepVGG-Based Single-Branch Inference Module*

In this sub-section, to make the whole multimodal emotion recognition system better adaptive to future videos in the wild with high resolution and long length, we attempt to deal with the problem of low inference efficiency due to the complex structures of common deep learning models and meanwhile we aim to guarantee the performance.

For multimodal emotion recognition tasks, we pioneer the use of the multi-branch feature learning and single-branch inference structure proposed in the latest RepVGG technology [39]. In Fig. 3, we show the training and the inference models adopted for the visual modality, whose structure is based on $3 \times 3$ convolution, $1 \times 1$ convolution, Identity, and ReLU activation layers [40]. In the left panel of this figure, we can see that spatial-visual feature is learned mainly through multi-branch and multi-kernel convolutions. As shown in Fig. 3, for the input image frame, we use three branches to extract multi-dimensional features, including the $3 \times 3$ convolution branch, the $1 \times 1$ convolution branch, and the Identify branch. Then the middle features are fused as the input of the ReLU layer [40] to obtain the final result.

The right panel of Fig. 3 shows the inference procedure, instead of the original clunky multi-branch structure, we choose a purer single-branch structure. And the middle panel shows the details of the process of parameter migration from training models to inference models, in which a three-channel input is taken as an example. Specifically, among the three branches in training models, only the $3 \times 3$ convolution branch is retained while the $1 \times 1$ convolution kernel is transformed to $3 \times 3$ convolution by zero padding. It is worth noting that each convolution here contains a BN layer [41].

The original RepVGG used a model architecture similar to ResNet in order to benchmark the SOTA model. However, the problem we face is that of improving the processing efficiency of video image frames when dealing with a lot of high-definition videos. Thus, we use the multi-branch training and single-branch prediction ideas of RepVGG to achieve this target. And we hope to improve inference efficiency with a simpler structure. Hence, we combine a $3 \times 3$ convolution, a $1 \times 1$ convolution and Identity as one layer, and finally use such six layers as the visual module structure for the target task.

*C. The Fully Video-to-Emotion System*

Although some video sentiment analysis also performs preprocessing operations, they often choose to store modal information files after video processing. Then the model calls the modal information file for inference analysis. In this process, the memory and efficiency waste generated by storage, generation and invocation, we think, unnecessary. We want to solve the "last mile" problem by decreasing the unnecessary computational storage waste. Thus, we designed a "fully video to emotion system" (FV2ES), which directly interfaces the preprocessed modal information with the V2EM input in order to reduce the waste of previous systems.

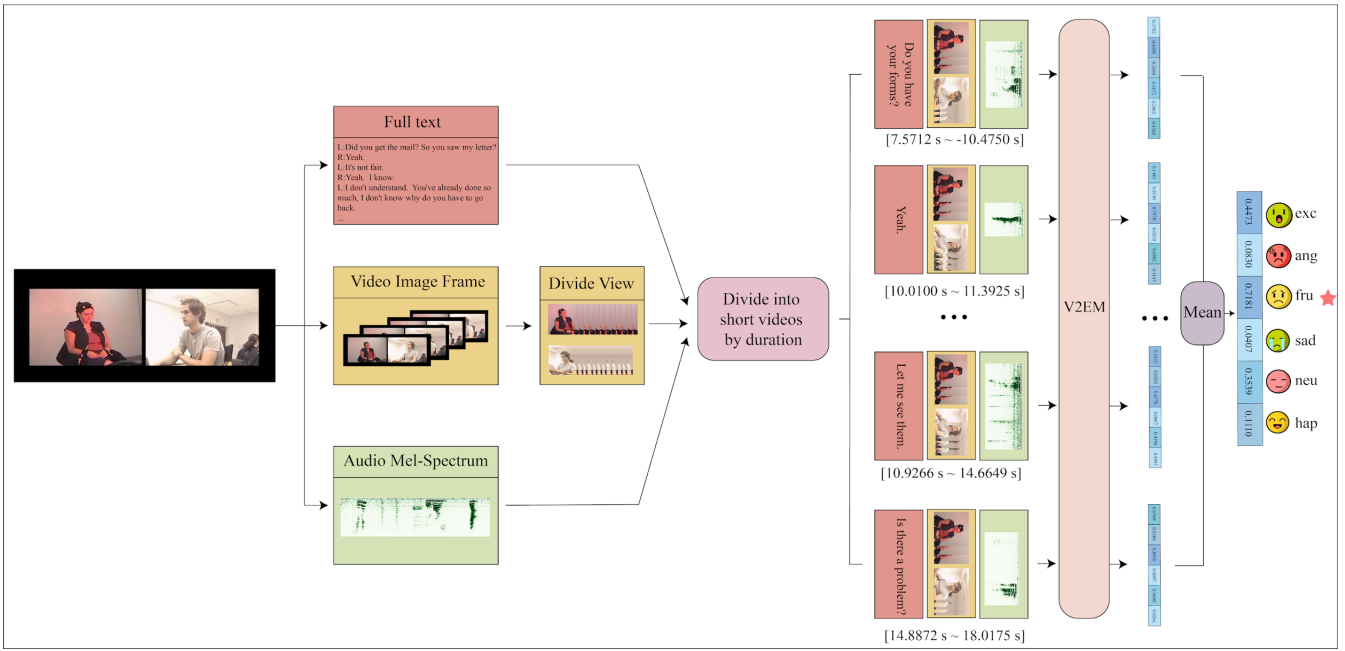

Fig. 4. The architecture of FV2ES.

In this subsection, we take the first step to think outside of the existing academic research on multimodal video emotion recognition. Fig. 4 illustrates the data flow of FV2ES, in which the whole process including data uploading, features extraction, and the final emotion prediction is coherent. In FV2ES, long videos are allowed to be taken as the input, and the complete dialogue texts and the audio Mel-spectrogram of the input videos are obtained firstly. At the same time, the image frames need to be preprocessed. Take the videos in the IEMOCAP dataset, for instance. Here the two-person dialogue videos are divided into one-person frames according to the characters. Long videos are too verbose and cause computational overhead. To solve this problem, we divide the whole video into several short video segments. At the same time, we use the timeline to align the data of the three modalities. We select the text, spectrum, and image frames in the same period and take these directly as input to V2EM. Then we can obtain the emotion prediction of this period through V2EM. Finally, the emotional prediction values of multiple short videos are taken as the emotional prediction result of the input video.

Based on the Flask technology, we build FV2ES which is expected to enhance the industrial application value of multimodal video emotion analysis. Users can upload the videos to be analyzed at the front end of the platform while these modality data will be computed at the back end. The obtained emotional scores and the final prediction will be printed on our designed interface.

## IV. EXPERIMENTS AND ANALYSIS

### A. Experimental Environment and Datasets

The Jiutian artificial intelligence platform is chosen as the experimental platform for this experiment. This is an artificial intelligence innovation platform independently developed by China Mobile, providing open AI services from infrastructure to core capabilities. The experiments in this paper rely on the Tesla V100S-PCIE-32GB GPU and PyTorch v1.8.0.

There are two public datasets used in our experiments, including the IEMOCAP and the CMU-MOSEI datasets. The IEMOCAP dataset consists of multimodal data of three modalities of video, audio, and text transcription. We select six main categories from the original emotions: anger, happiness, excitement, sadness, frustration, and neutral. And to create a new split for the dataset, we randomly assign 70%, 10%, and 20% of the data to the training, validation, and test sets respectively.

The CMU-MOSEI dataset also consists of multimodal data of three modalities of vision, audio, and text. Six kinds of labels including happiness, sadness, anger, fear, disgust, and surprise were annotated for the videos. And the dataset contains 250 topics, 3837 videos, 23453 sentences, 1000 narrators, and the total duration reaches 65 hours.

In subsequent experiments, the same dataset partitioning method was also used for all the baselines.

### B. Implementation Details

As to evaluation indicators, we use Accuracy and F1 score to evaluate our proposed model on the IEMOCAP dataset, while we take weighted accuracy ($W_{Acc.}$) instead of standard accuracy on the CMU-MOSEI dataset. The formula for weighted accuracy is:

$$W_{Acc.} = \frac{TP \times N/P + TN}{2N} \quad (4)$$

### TABLE II
PREDICTION PERFORMANCE OF V2EM AND EXISTING MODELS ON THE IEMOCAP DATASET. $W_{Acc}$ AND F1 ARE THE AVERAGE VALUES OF THE SIX EMOTIONS. SPEED IS THE TOTAL TIME (S) FOR TESTING 1481 SAMPLES.

| Model | $W_{Acc}$ | F1 | Speed |
|---|---|---|---|
| LF-LSTM (Baseline) | 0.718 | 0.495 | - |
| LF-TRANS (Baseline) | 0.788 | 0.503 | - |
| EmoEmbs (year 2020) [39] | 0.720 | 0.498 | - |
| MulT (year 2019) [23] | 0.776 | **0.569** | - |
| FE2E (year 2021) [5] | 0.819 | 0.558 | 388.24 |
| V2EM (ours) | **0.829** | 0.564 | **186.55** |

### TABLE III
PREDICTION PERFORMANCE OF V2EM AND EXISTING MODELS ON THE CMU-MOSEI DATASET. $W_{Acc}$ AND F1 ARE THE AVERAGE VALUES OF THE SIX EMOTIONS. SPEED IS THE TOTAL TIME (S) FOR TESTING 4188 SAMPLES.

| Model | $W_{Acc}$ | F1 | Speed |
|---|---|---|---|
| LF-LSTM (Baseline) | 0.631 | 0.433 | - |
| LF-TRANS (Baseline) | 0.641 | 0.444 | - |
| EmoEmbs (year 2020) [39] | 0.642 | 0.442 | - |
| MulT (year 2019) [23] | 0.654 | **0.452** | - |
| FE2E (year 2021) [5] | 0.691 | 0.446 | 376.36 |
| V2EM (ours) | **0.699** | 0.448 | **295.76** |

### TABLE IV
RESULTS OF EACH MODEL TRAINING ON THE IEMOCAP DATASET. IN 30 EPOCHS, 1481 TEST SAMPLES IN IEMOCAP ARE TESTED WITH A BATCH SIZE OF 8. THE BASE V+A+T MODEL USED THE VGG16 MODEL TO PROCESS VISUAL AND AUDITORY DATA, AND IT USED ALERT TO EXTRACT TEXT FEATURES. IN THE V+A(OURS)+T MODEL, THE HIERARCHICAL-ATTENTION SPECTRUM COMPUTING MODULE IS USED TO EXTRACT AUDIO FEATURES, AND IN THE V(OURS)+A+T MODEL, THE REPVGG-BASED SINGLE-BRANCH INFERENCE MODULE IS USED TO EXTRACT VISUAL FEATURES. $W_{Acc}$ AND F1 ARE THE EVALUATION INDICATORS.

| Model | $W_{Acc}$ | | F1 | |
|---|---|---|---|---|
| | Max | Average | max | average |
| **V+A+T** (Baseline) | 0.8625 | 0.8192 | 0.6170 | 0.5583 |
| **V+A(Ours)+T** | 0.8519 | 0.8223 | 0.6035 | 0.5603 |
| **V(Ours)+A+T** | 0.8525 | 0.8006 | 0.5859 | 0.5337 |
| **V(Ours)+A(Ours)+T** | 0.8584 | **0.8292** | 0.5999 | **0.5644** |

### TABLE V
RESULTS OF EACH MODEL TRAINING ON THE CMU-MOSEI DATASET. IN 30 EPOCHS, 4188 TEST SAMPLES IN CMU-MOSEI ARE TESTED WITH A BATCH SIZE OF 8.

| Model | $W_{Acc}$ | | F1 | |
|---|---|---|---|---|
| | max | average | max | average |
| **V+A+T** (Baseline) | 0.729 | 0.6913 | 0.4619 | 0.4460 |
| **V+A(Ours)+T** | 0.7224 | **0.6992** | 0.4638 | 0.4428 |
| **V(Ours)+A+T** | 0.7222 | 0.6811 | 0.465 | 0.4462 |
| **V(Ours)+A(Ours)+T** | 0.734 | 0.6987 | 0.4748 | **0.4476** |

where $P$ represents total positives, $TP$ is the number of true positives, $N$ is total negatives, and $TN$ represents the number of true negatives.

In this paper, we use the following five state-of-the-art models in previous work as the baselines.

- **LF-LSTM**: Late fusion using LSTM.
- **LSTMLF-TRANS**: Late fusion using Transformer.
- **EmoEmbs** [42]: Using a multimodal transferable model. And pre-trained word embeddings were used to represent textual features. And two mapping functions were built to transfer the embeddings into visual and auditory modalities. Then calculate the distance between the predicted and the target value for each modality, and make predictions based on the distance.
- **MulT** [23]: The structure of the Cross-modal Transformer was used to construct the relationship between different modalities. After obtaining the multimodal fusion information, three sets of features were joined for prediction.
- **FE2E** [5]: Extract visual and audio features using VGG16 and text features using Alert.

Moreover, the Adam optimizer [43] is used to train each model. And the binary cross-entropy loss is adopted as the loss function. The hyperparameters such as the learning rate, the epoch and the batch size are set to 4.5e-6, 30 and 8 in our experiments.

*C. Experimental Results and Analysis*

*1) Recognition Performance*

This section demonstrates the advantages of the performance of our proposed V2EM for multimodal video emotion recognition.

**Comparative Experiments.** In Table II, we show the results of V2EM on the IEMOCAP dataset.

Compared to baselines, the weighted accuracy of V2EM is outstanding (15.46% improved compared to LF-LSTM, 5.20% improved compared to LF-TRANS, 10.9% improved compared to EmoEmbs, and 5.30% improved compared to MulT). Such excellent results confirm the superiority of the V2EM structure. Since FE2E performs best in the existing SOTA models and it used the same text processing method as V2EM, we choose FE2E as the reference model to observe the effects of V2EM in the 30 epochs training process. As shown in Fig. 5, during 30 epochs, the performance of V2EM is significantly better than that of FE2E (**1.22%** improved), and the good performance is maintained stably. We further evaluate the results of V2EM on the CMU-MOSEI dataset as shown in Table III and Fig. 5. We observe similar trends on this dataset. (10.78% improved compared to LF-LSTM, 9.05% improved compared to LF-TRANS, 5.70% improved compared to EmoEmbs, 4.50% improved compared to MulT and 1.16% improved compared to the best SOTA model FE2E [5]).

**Ablation Experiments.** To better explore the effect of the hierarchical-attention spectrum computing module and the RepVGG-based single-branch inference module on multimodal video emotion prediction, we conduct ablation experiments, in which the VGG16 model is used to process the visual and auditory data, and the Alert model is adopted to extract textual features. Then we use the hierarchical-attention spectrum computing module and the RepVGG-based single-branch inference module respectively on the basis of the basic model to observe the prediction performance.

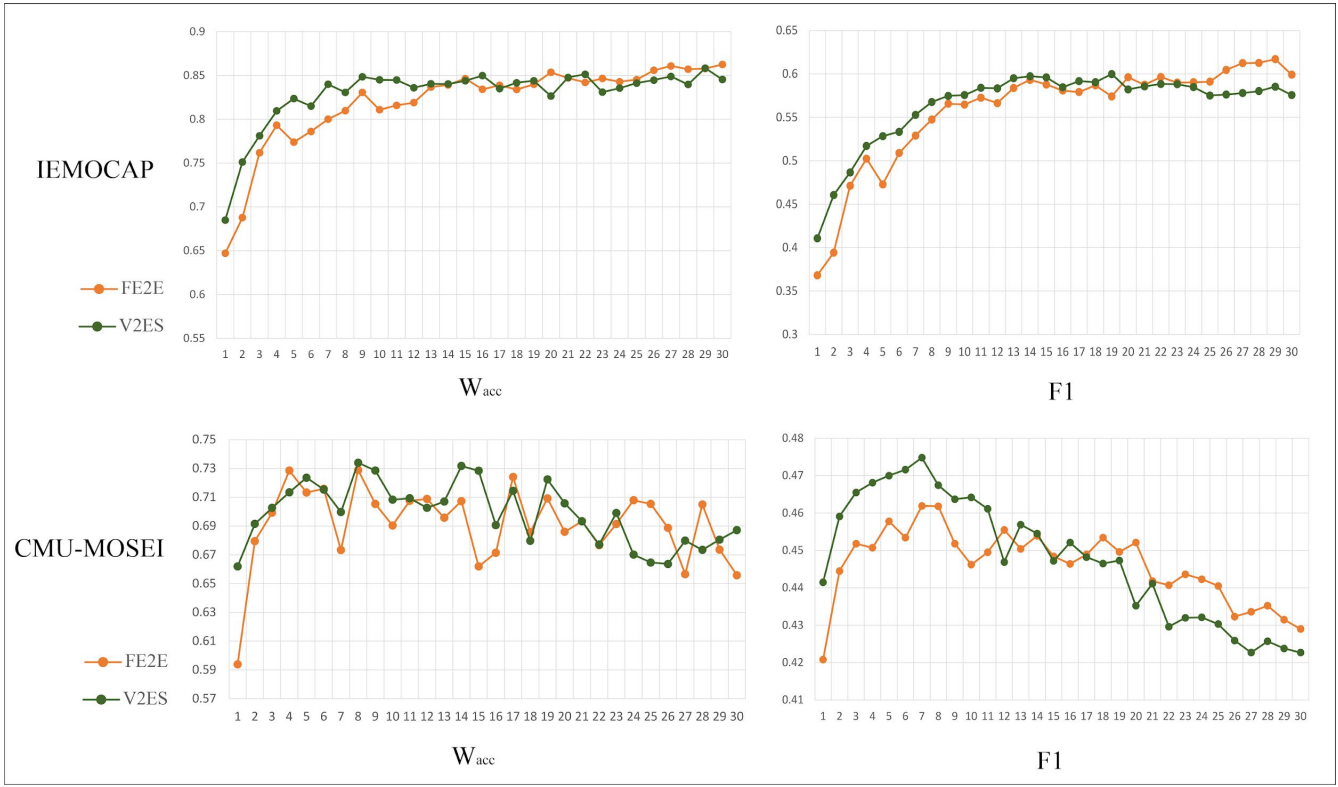

Fig 5. The performance changes of FE2E (year 2021, the best SOTA model), V2EM (ours) on the IEMOCAP dataset and the CMU-MOSEI dataset during 30 epochs. In each table, the horizontal axis represents the number of epochs, and the vertical axis represents the $W_{Acc}$ value (F1-score)

TABLE VI

**THE INFERENCE SPEED OF EACH MODEL ON THE IEMOCAP DATASET.** IN 30 EPOCHS, 1481 TEST SAMPLES ARE TESTED WITH A BATCH SIZE OF 8. SPEED IS THE TOTAL TIME (S) FOR TESTING 1481 SAMPLES, IMPROVED IS THE RELATIVE IMPROVEMENT OF SPEED COMPARED TO THE BASELINES.

| Model | Speed | Improved |
|---|---|---|
| V+A+T (Baseline) | 388.24 | - |
| V+A(Ours)+T | 198.50 | 48.87% |
| V(Ours)+A+T | 249.42 | 35.76% |
| V(Ours)+A(Ours)+T | **186.55** | **51.95%** |

TABLE VII

**THE INFERENCE SPEED OF EACH MODEL ON THE CMU-MOSEI DATASET.** IN 30 EPOCHS, 4188 TEST SAMPLES ARE TESTED WITH A BATCH SIZE OF 8. SPEED IS THE TOTAL TIME (S) FOR TESTING 4188 SAMPLES, IMPROVED IS THE RELATIVE IMPROVEMENT OF SPEED COMPARED TO THE BASELINES.

| Model | Speed | Upgrade |
|---|---|---|
| V+A+T (Baseline) | 376.36 | - |
| V+A(Ours)+T | 308.05 | 18.03% |
| V(Ours)+A+T | 297.62 | 20.92% |
| V(Ours)+A(Ours)+T | **295.76** | **21.42%** |

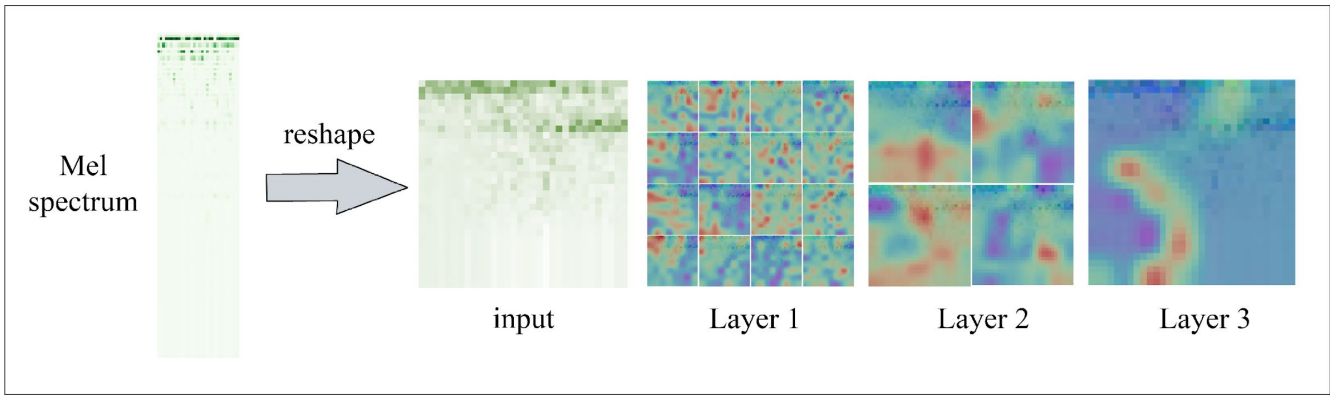

Fig. 6. The audio attention maps. We take the first audio clip of Ses01M_script03_2 in the IEMOCAP dataset as the example. And we show the attention map results of each layer. Warm areas represent the focused regions of the attention layers.

TABLE VIII

THE NUMBER OF PARAMETERS IN EACH MODALITY OF FE2E AND V2EM.

| Modal | FE2E | V2EM |
|---|---|---|
| T | 11.69M | 11.69M |
| V | 10.71M | **7.72M** |
| A | 14.38M | 101.99M |

As shown in Table IV and Table V, the introduction of the audio module effectively improves the performance on these two public datasets. On the IEMOCAP dataset, $W_{Acc}$ in the V+A(Ours)+T model equals 0.8223, which is **0.38%** improved compared to the baseline, and the F1 score is **0.36%** improved compared to the baseline. Similarly, on the CMU-MOSEI dataset, the adoption of the audio module makes $W_{Acc}$ **1.14%** improved. At the same time, the performance of the V+A(Ours)+T model on both datasets is significantly better than that of V(Ours)+A+T, indicating that the audio modality has a greater contribution to the overall multimodal emotion recognition results. Here, it can be seen that the effect of V(Ours)+A+T is not as good as that of baselines. We believe this is due to the relative simplicity of our six-layers visual structure compared to the VGG-16 structure used by the baseline. Thus, there is less performance improvement. However, our original intention of designing the RepVGG-based single-branch inference module is to bring efficiency improvement**s**, which is demonstrated in the subsequent analysis. Thus, its lack of performance is something we can tolerate. Therefore, the above experiments show that the proposed V2EM using the hierarchical-attention spectrum computing module can solve the problem that the contribution of the audio modality is relatively low in the existing multimodal models.

To make further analysis of the reason for the good performance of the hierarchical-attention spectrum computing module proposed in section III-A, we show a visualization of audio attention maps in Fig. 6. In this figure, we take the first audio clip of Ses01M_script03_2 in the IEMOCAP dataset as the example. The audio attention maps of each layer are extracted through the self-attention module of three layers. On the far left is the original Mel spectrogram. Reshape Mel Spectrum to get the input of the audio module, as shown in the "Input". After being processed by the first Transformer layer of the audio module, the 16 attention maps shown in "Layer 1" are obtained. The warm areas represent the focused regions of the attention layers. We can observe that the features of the hierarchical-attention spectrum computing module focus on more subtle internal relationships of audio features in early attention layers. Then, after Aggregation and the second Transformer layer, 4 audio attention maps are obtained as shown in "Layer 2". Its warm area becomes larger. At the same time, as shown in "Layer 3", the warm area in the attention map obtained after the third Transformer layer is larger. It can be seen that the information on the internal features of the final fusion sound spectrum can be obtained after hierarchical extraction. Therefore, we can say that the proposed audio module can extract more fine-grained, locally high and low

TABLE IX

DUE TO LIMITED GPU MEMORY, WE HAVE TRIED OUR BEST TO RANDOMLY SELECT TEN TESTING SAMPLES FROM THE IEMOCAP DATASET TO TEST THE PREDICTION EFFICIENCY OF PRE+FE2E, PRE+V2EM AND FV2ES. PRE+FE2E, PRE+V2EM MEANS THAT THE VIDEO PREPROCESSING STORES EACH MODAL INFORMATION, AND THE MODEL CALLS THE PREPROCESSING INFORMATION FOR PREDICTION. IT SHOWS THE TOTAL TIME (S) FOR TESTING. IMPROVED IS FV2ES'S IMPROVEMENT COMPARED TO PRE+V2EM.

| Testing samples | Pre+FE2E | Pre+V2EM | FV2ES | Improved |
|---|---|---|---|---|
| Ses01F_impro02 | 129.56 | 121.53 | 37.66 | 69.01% |
| Ses01M_script03_2 | 175.46 | 165.01 | 58.02 | 64.84% |
| Ses02F_impro02 | 142.51 | 128.25 | 52.09 | 59.38% |
| Ses02M_script01_2 | 99.13 | 93.11 | 50.19 | 46.10% |
| Ses03F_impro07 | 135.08 | 116.37 | 68.74 | 40.93% |
| Ses03F_script01_3 | 240.39 | 222.86 | 50.24 | 77.46% |
| Ses04F_impro01 | 129.63 | 123.74 | 55.46 | 55.18% |
| Ses04M_script02_2 | 172.42 | 171.89 | 55.59 | 67.66% |
| Ses05F_impro03 | 171.70 | 158.36 | 67.48 | 57.39% |
| Ses05M_impro07 | 176.55 | 167.67 | 47.40 | 71.73% |
| *Average* | **157.24** | **146.88** | **54.29** | **63.04%** |

audio spectral features, thus leading to the enhancement of the acoustic contribution to the overall multimodal emotion recognition performance.

*2) The Inference Efficiency of V2EM*

**Comparative Experiments.** In Table II and Table III, we show the prediction speed of V2EM on the IEMOCAP dataset and the CMU-MOSEI dataset. V2EM is the proposed end-to-end model for multimodal video-to-emotion prediction. And FE2E is the best end-to-end prediction model in the SOTAs. From these two tables, we can find that the computing efficiency of V2EM is outstanding (On the IEMOCAP dataset, **51.95%** improved compared to FE2E while **21.42%** improved on the CMU-MOSEI dataset).

**Ablation Experiments.** As shown in Table VI and Table VII, the application of the RepVGG-based single-branch inference module effectively improves the inference efficiency on the two datasets. Specifically, on the IEMOCAP dataset, V(Ours)+A+T takes 249.42 seconds to test 1481 samples, and the improvement compared to the baseline is **35.76%**. Similarly, on the CMU-MOSEI dataset, V(Ours)+A+T takes 297.62 seconds to test 4188 samples, and the improvement compared to the baseline is **20.92%**. Overall, the processing of the visual branch greatly enhanced the model inference efficiency so that it can help cope with today's videos with high resolution and long length. The main reason is due to the function of the single-branch inference structure in the vision processing module proposed in section III-B.

In Table VIII, the numbers of parameters involved in each modality of the SOTA FE2E model and our proposed V2EM model are given. It shows that for the visual modality, there are significantly fever parameters due to the adopted single-branch inference structure. And this can help us explain the good effect of the reduced computational complexity and the improved inference efficiency.

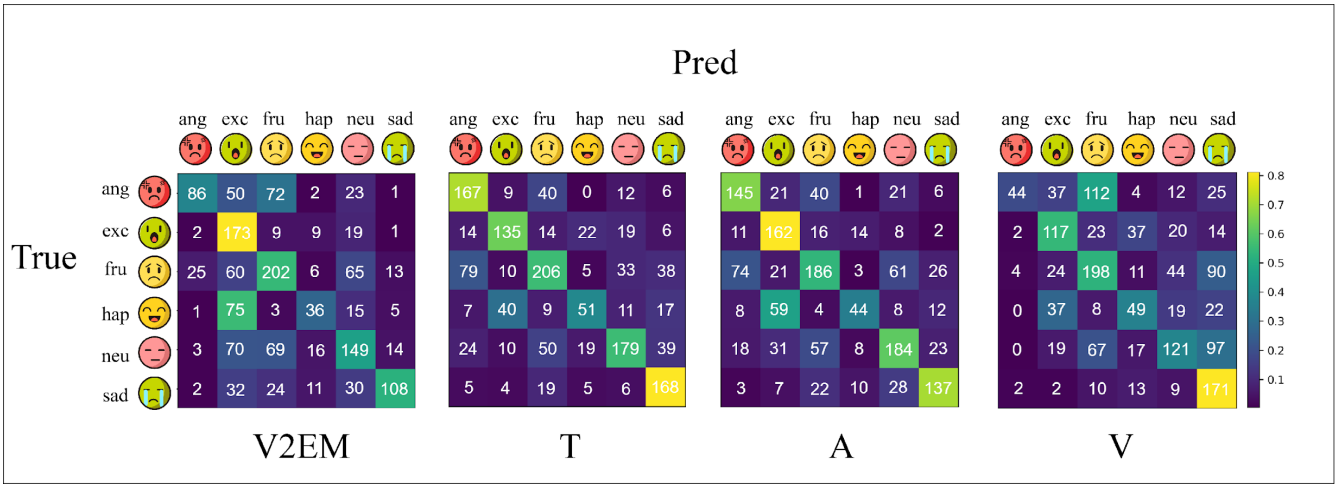

Fig. 7. Confusion **matrices** for V2EM and the prediction results for each of the three modalities on the IEMOCAP dataset. There are 1481 test samples with six emotion categories: anger, excitement, frustration, happiness, neutral, and sadness.

*3) The Inference Efficiency of FV2ES*

On the basis of V2EM, the data pre-processing and the multimodal end-to-end learning model are integrated to achieve FV2ES, in which we dispense with the operations of saving and uploading the preprocessed data of various modalities, aiming to reduce the need for unnecessary data storage space.

In Table IX, we compare the inference efficiency results of Pre+FE2E (using independent video pre-processing in advance and the baseline model for emotion prediction), Pre+V2EM (using independent video pre-processing plus our proposed V2EM model), and the designed FV2ES in this paper. It can be seen in Table IX that FV2ES greatly enhances the prediction efficiency (**63.04%** improved compared to Pre+V2EM). **Combined with the comparison performances in TABLE II and TABLE III, we can conclude that the proposed FV2ES brings fast yet effective video emotion recognition inference.**

## V. Conclusion and Discussion

In this paper, we transfer the success of VIT to the audio modality for the multimodal emotion recognition task. **Taking layered audio spectrum information into consideration** can effectively improve the results of emotion classification, and meanwhile it can also promote the contribution of the acoustic modality compared to the visual and textual modalities for the overall multimodal task. **In addition, we introduce the RepVGG-based single-branch inference module** for visual frames. In this module, the "big and whole" instead of "small and fragmented" reasoning structure brings much higher computing efficiency and requires less storage space. And therefore, we expect this vision module to help solve the computational and storage problem caused by a large amount of long-length and high-resolution videos in the 5G and self-media era. **Moreover, we designed a fully multimodal video-to-emotion system** in which the data pre-processing and the multimodal end-to-end learning model are integrated. This operation leads to the further enhancement of the inference efficiency. **Above all, the proposed FV2ES in this paper for fast yet effective video emotion recognition inference is valuable in academic study and also applicable for practical industry.**

However, FV2ES still has some shortcomings. Fig. 7 illustrates the results of the confusion matrix. We can observe that the method proposed in this paper has the best recognition performance on emotion "excitement". The main reason is that the excitement is more often expressed in speech and intonation, and the proposed audio module functions well in extracting fine-grained acoustic information (as shown in the confusion matrix A modality). But for similar emotional expressions like happy and neutral, people express subtle differences in expressions, dialogues, and voices, and our model does not perform well in discriminating these emotions.

At the same time, the video scenes in the public IEMOCAP and CMU-MOSEI datasets used in this paper are relatively pure instead of in the wild. Specifically, the environment in these videos is quiet and there is almost no noise. Furthermore, there are easily recognizable core characters in the frames. Such clean datasets are still far from the real world. In the future, we will optimize V2EM to improve the performance and migrate FV2ES to short video data in social networks. And we are going to take the visual processing of complex backgrounds, noisy audio, and random text processing into account, so as to promote more links between the multimodal video emotion analysis academic research and industry application.

## VI. Acknowledgments

This work was supported by the National Natural Science Foundation of China under Grant No. 62271455, the project funded by China Postdoctoral Science Foundation (ZW21099), and the Fundamental Research Funds for the Central Universities (CUC220F003 & CUC21GZ012).

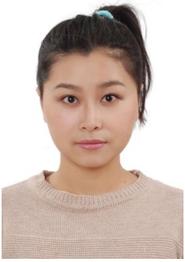

**Qinglan Wei** (Member, IEEE) is currently a Post-doc researcher with the School of Data Science and Intelligent Media, Communication University of China. She was a visiting scholar at the Language Technologies Institute, Carnegie Mellon University, USA. She received the Ph.D. degree in computer science from the College of Artificial Intelligence, Beijing Normal University, China. She has published a series of articles in academic journals and conferences. She has twice received the Second Runner-up position of the Group Emotion Recognition Sub-challenge sponsored by ICMI, ACM. Her main research interests include machine learning, computer vision, and emotion analysis.

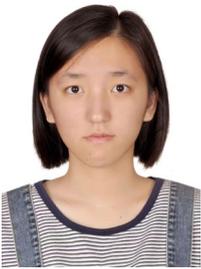

**Xuling Huang** is currently studying data science and big data technology at Communication University of China. Her research interests include multimodal emotion recognition, deep learning and image processing.

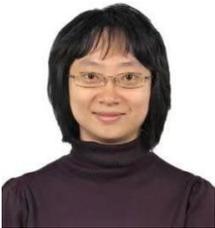

**Yuan Zhang** received the B.Sc. and M.Sc. degrees in electronic engineering from Communication University of China (CUC), Beijing, in 1995 and in 1998, respectively, and the Ph.D. degree in computer software and theory from the Graduate School of the Chinese Academy of Sciences, in 2007. She was a Visiting Scholar with the Department of Electrical and Computer Engineering, University of California, San Diego, USA, from September 2010 to August 2011. She is currently a Full Professor, the Vice Director of State Key Laboratory of Media Convergence and Communication, Communication University of China, Beijing.